\documentclass{article} % For LaTeX2e
\usepackage{iclr2027_conference,times}

\usepackage{amsmath,amsfonts,bm}

\def\eqref#1{equation~\ref{#1}}
\def\1{\bm{1}}

\DeclareMathAlphabet{\mathsfit}{\encodingdefault}{\sfdefault}{m}{sl}
\SetMathAlphabet{\mathsfit}{bold}{\encodingdefault}{\sfdefault}{bx}{n}

\usepackage{hyperref}
\usepackage{url}
\usepackage{amssymb}
\usepackage{siunitx}
\usepackage{booktabs}
\usepackage{multirow}
\usepackage{graphicx}
\usepackage{float}
\usepackage[table]{xcolor}

\title{RepProbe: Can Video Language Models Count Repeated Actions? A Diagnostic Benchmark}

\author{Mahsa Khoshnoodi, Sarah Adel Bargal  \\
Department of Computer Science, Georgetown University\\
}

\iclrfinalcopy % Uncomment for camera-ready version, but NOT for submission.
\begin{document}

\maketitle
\lhead{Preprint} 
\begin{abstract}

Repetition counting requires identifying and enumerating action cycles over
time. While state-of-the-art video language models describe actions fluently,
they fail to reliably count how many times an action repeats. To investigate
this limitation, we introduce \textbf{RepProbe}, a benchmark of $42$K real and
synthetic video clips containing exactly $k \in [1, 20]$ complete repetitions,
with controls for clip position, temporal coverage, and duration. These
controls allow us to study the counting failure directly and locate where in
the model it arises. Counting accuracy collapses at high counts across the
video language models we evaluate, and controlled experiments show this
collapse is not explained by frame sampling alone. We compare model outputs
with linear probes trained on the final-layer representations used to
generate those outputs. We find a consistent representation-to-output
dissociation: substantial count information remains linearly decodable even
when models fail to report the correct count. Further analysis shows this failure arises primarily at the output stage, where answers concentrate on a small set of preferred numbers rather than the available count information.
The dissociation holds on both real and synthetic videos and is not explained by clip duration, position, or scene content. These results indicate that
models encode repetition structure but do not reliably report it in their outputs.
\end{abstract}

%Sarah: The second half of the abstract reads like a results section

%To locate the failure, we look inside the two strongest models (Cosmos-Reason2 and Qwen3-VL),
% comparing the model's answer with a linear probe trained to read the count from
% the same final-layer representation. The probe recovers much of the count where
% the model's answer fails at high $k$: the count is present in the representation
% but not reliably reported. We trace the failure to the output stage, where the
% model defaults to a few common numbers rather than reporting the count it holds.
% Removing this bias improves accuracy across four model families, while amplifying
% the count itself does not, showing the count is present but overridden at the
% output. Finally, to test whether this reflects the count itself or a scene-level
% cue, we build a 10K synthetic set of periodic motion with no objects or scene,
% where the count is defined only by repetition over time. There the probe still
% recovers the count (MAE $0.62$ and $0.59$) while the model's answer fails
% ($7.75$ and $7.77$), showing the decodable signal is the temporal repetition
% structure itself.

\section{Introduction}
\label{sec:intro}

Counting how many times an action repeats is a basic temporal skill. It
requires identifying discrete action instances, recognizing the cycles of a
periodic action, and accumulating these repetitions as they unfold over time.
Unlike static counting, these repetitions are distributed across a sequence,
so no single moment contains the answer. Human cognition treats this capacity
as distinct: the number of sequential events is represented separately from
other temporal properties such as duration \citep{white1963temporal}, and
parietal populations encode how many times an action is repeated
independently of the action's identity \citep{sawamura2020parietal}. Counting
therefore differs from recognizing an action. Recognition, even when it
requires temporal information, is a categorical judgment about presence or
identity; counting tracks each occurrence and accumulates them into an exact
quantity. This makes repetition counting a precise test of whether a model
tracks temporal structure, and it is also required in practice, for instance
in exercise and rehabilitation monitoring, where approximate counts are of
limited use \citep{dwibedi2020counting,hu2022transrac}.

Video Language Models (VLMs) handle this task poorly. They describe repetitive
actions fluently yet misreport how many times the actions occur, and their
error grows with the number of repetitions
\citep{yao2025countllmgeneralizablerepetitiveaction,li2026pushupbench}. The
difficulty does not lie in the task itself: specialist counters trained for
this purpose remain accurate at high counts, whereas general-purpose video
language models, despite their scale and broad video training, degrade as
repetitions accumulate. Existing evaluations of counting and related temporal
skills \citep{liu2024tempcompass,cores2024lost,patraucean2023perception},
however, measure only the final answer, and they do not separate repetition
count from factors that vary with it, such as clip duration, temporal
coverage, and source video. An incorrect answer does not imply that the
underlying information is absent: probing studies show that models can encode
information they do not express
\citep{alain2016understanding,burns2022discovering,azaria2023internal}. The
source of the failure therefore remains open: do these models fail to
represent the count, or represent it but fail to report it? The answer
matters, because a failure of representation and a failure of readout call
for different remedies.

To address this question, we introduce \textbf{RepProbe}, a diagnostic
benchmark in which every clip contains an exact, known number of complete
repetitions. In RepCount~\citep{hu2022transrac}, each video carries a single
count entangled with its duration and source. RepProbe-Real instead cuts clips
with exactly $k$ repetitions from each RepCount source video, controlling clip
position and source-level splits while auditing duration and temporal
coverage. RepProbe-Synthetic is a controlled set, balanced across counts, in
which only a single attribute changes periodically. We evaluate seven open
video language models and two specialist counters on RepProbe. To test whether
the count is absent from the model or present but not reported, we then
compare two readouts of the count from the same final-layer state of the
strongest generalists: the model's generated answer, and a linear probe on the
representation consumed by the language head. Finally, we test whether the
decodable count reflects the repetitions themselves, using duration and
clip-position controls on real video and controlled synthetic stimuli.

Our results show that the count is represented within the
model but not faithfully reported. Generalist
video language models count reasonably well at low counts but collapse at high
counts, where a specialist counter remains accurate, and this collapse is not
explained by frame sampling. Yet the count remains linearly decodable from the
final-layer state that the language head reads, even when the model's answer
is wrong. This dissociation also holds on synthetic stimuli, and it is not explained by duration or clip position. The generated answers instead concentrate on a small set
of preferred numbers, largely independent of the true count. These findings
indicate that the high-count failure of video language models arises in the
readout of the representation rather than in the representation itself. Our
analysis complements prior work showing that probes can recover temporal
information from vision-side features \citep{li2025temporal}: we localize the
gap to the final representation consumed by the language head, and show that
it widens as the count grows.

\paragraph{Our contributions are summarized as follows:}
\begin{itemize}
\item \textbf{RepProbe, a controlled benchmark for repetition counting.}
RepProbe pairs RepProbe-Real, $32$K clips constructed from the $1$K source
videos of RepCount \citep{hu2022transrac}, with RepProbe-Synthetic, $10$K
procedurally generated clips in which a single attribute changes periodically,
making $42$K clips in total, all with exact counts from $1$ to $20$. 

\item \textbf{A systematic evaluation revealing a high-count collapse.}
Across seven generalist open video language models and two specialist counters, generalist models
fall to at most $22.2\%$ OBO at high counts while specialist counters remain
accurate. This failure is obscured by overall accuracy and persists when the
number of frames per repetition is held fixed.

\item \textbf{Evidence that count is better represented than reported.}
A linear probe on the final-layer state recovers the count substantially more
accurately than the model's own answer, on both real and synthetic video,
while the answers concentrate on a small set of preferred numbers. This
locates the failure in the output pathway and identifies readout, rather than
perception, as the target for improving counting in video language models.
\end{itemize}

\section{Related Work}
\paragraph{Repetitive action counting.}
Repetitive action counting has been studied as a dedicated video understanding
problem. Early approaches estimate repetition periods from temporal
self-similarity (RepNet;~\citealt{dwibedi2020counting}), while later methods
address unknown and varying cycle lengths through context-aware,
scale-insensitive estimation \citep{zhang2020context}. Datasets progressed
from Countix \citep{dwibedi2020counting} and UCFRep
\citep{zhang2020context} to RepCount, which introduces variable-length videos
and fine-grained cycle annotations, together with density-map supervision for
repetition-period estimation \citep{hu2022transrac}. Recent specialist
counters include Every Shot Counts (ESCounts), which uses exemplar-based
visual correspondence across repetitions \citep{sinha2024every}, PAMS, a
self-supervised method based on period-adaptive multi-scale consistency
\citep{Gao_2026_CVPR}, and multimodal methods that combine appearance and pose
to model repetitive motion under background variation
\citep{Wang_2026_CVPR}. These methods are trained for counting and optimize
counting accuracy. Together, they establish repetition counting as a
well-defined video understanding target with a measurable ground-truth
quantity, making it a controlled setting for testing whether general-purpose
video language models encode the information needed for counting.

\paragraph{Counting and temporal reasoning in vision-language models.}
Counting remains challenging for vision-language models even on static images
\citep{acharya2018tallyqaansweringcomplexcounting,paiss2023teaching}, and
recent perception benchmarks continue to identify failures on basic visual
counting and spatial tasks \citep{rahmanzadehgervi2024vision,fu2024blink}.
In video, counting becomes temporal, since events must be accumulated across
frames. Video benchmarks such as MVBench include temporal action-understanding
tasks \citep{li2024mvbench}, while CountLLM
\citep{yao2025countllmgeneralizablerepetitiveaction} and the
exercise-counting benchmark PushupBench \citep{li2026pushupbench} report
substantial difficulties for general-purpose video language models on repetition
counting. More broadly, benchmarks evaluate fine-grained temporal
\citep{li2024mvbench,liu2024tempcompass}, long-form
\citep{wu2024longvideobench}, and visual-temporal reasoning
\citep{shangguan2025tomato}, but existing evaluations can still admit
solutions based on a few frames or without preserving temporal order.
RepProbe makes repetition count the controlled temporal variable and audits
alternative cues such as temporal position, sampling, duration, and source.
Whereas these benchmarks primarily measure failure at the output, RepProbe
localizes it by testing whether count information is absent from the
representation or present but not surfaced by the model's native output.

\paragraph{Representation probing.}
Linear probing provides a standard way to measure whether task-relevant
information is recoverable from frozen representations
\citep{alain2016understanding,belinkov2022probing}, and has shown that LLMs
linearly encode structured quantities such as space, time, truth, and numeric
value \citep{gurnee2024language,marks2023geometry,heinzerling-inui-2024-monotonic,
zhu2025language}. Prior work has also shown that probes can recover
information that is not reliably expressed in model outputs
\citep{burns2022discovering,azaria2023internal,orgad2025llms}, with related
work finding similar gaps in vision-language models
\citep{zhao2024first}. Interpretability studies of video LLMs further examine
where visual and temporal information is integrated across layers
\citep{kim2026map,shi2025causalitymatterstemporalinformation}. However, decodability does not
imply that the model uses the information
\citep{ravichander2021probing,elazar2021amnesic}. We therefore do not treat
probe accuracy as evidence of capability: the probe tests whether count
information is decodable from a representation, which we compare directly
with the information surfaced by the model's native language head. This
distinction is particularly relevant to recent work showing that temporal
information can be decoded from video representations even when Video LLMs
perform poorly on corresponding temporal reasoning tasks
\citep{li2025temporal}. Whereas that work probes an earlier vision-side
representation, we examine the final-layer vector directly consumed by the
language head.

\section{The RepProbe Benchmark}
\label{sec:dataset}

Existing repetition-counting benchmarks such as RepCount~\citep{hu2022transrac}
were designed to train and evaluate specialist counters, and they are not
suited to diagnosing why a model miscounts. Each RepCount video carries a
single count determined by its full length, so the count cannot be varied
independently of the video itself: comparing different counts means comparing
different videos, which also differ in duration, action, and speed. A model's
error on such a benchmark therefore cannot be attributed to the count rather
than to these correlated factors. The distribution of counts across videos is
also uneven, so some counts are represented by few videos. Diagnosing counting
failures requires varying the count while holding other factors fixed, which
motivates the construction of RepProbe.

\textbf{RepProbe} addresses these limitations by cutting clips with an exact,
known number of repetitions, so that the count can vary while the source video
and action stay fixed. It contains $42{,}233$ clips in two complementary parts
that share the same task, count range ($k=1$--$20$), and metrics.
\textbf{RepProbe-Real} ($32{,}233$ clips) is built from RepCount videos and
retains their natural visual variability, while giving precise control over
the number of repetitions in each clip and where the clip is taken from its
source. \textbf{RepProbe-Synthetic} ($10{,}000$ clips) is procedurally generated in a controlled setting where a single attribute changes periodically, so that counting can be tested when periodicity is the only available signal. In both parts, each clip
contains exactly $k$ complete periods, so counting difficulty can be evaluated
systematically as a function of $k$.

\subsection{RepProbe-Real}
\label{sec:dataset-real}

\paragraph{Construction.}
We build RepProbe-Real from exercise videos offering a controlled
setting for studying temporal counting, not because exercise counting is
uniquely difficult. Exercise repetitions are well defined: each repetition is
a complete cycle of a single action, typically beginning and ending in a rest
pose, so repetition boundaries are unambiguous and can be annotated precisely.
RepCount~\citep{hu2022transrac} provides repetition-boundary annotations that allow us to segment videos into clips containing exactly a specified number of complete repetitions. The synthetic set
(Section~\ref{sec:dataset-synthetic}) complements this choice by testing
whether our findings extend beyond exercise content.

\begin{figure}[t]
\centering
\includegraphics[width=\linewidth]{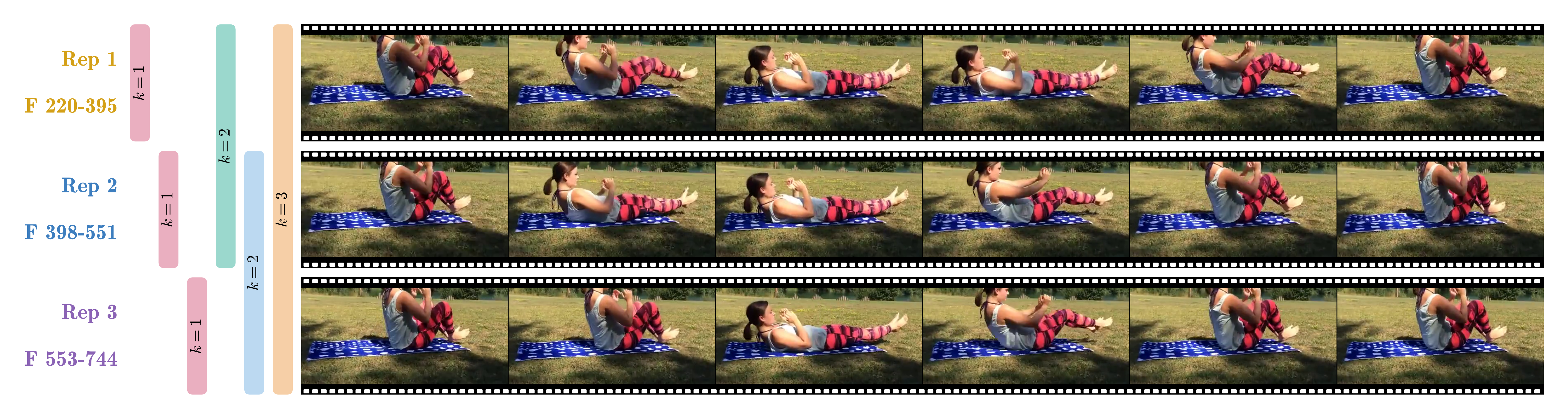}
\caption{\textbf{RepProbe-Real construction.} Rest-pose timestamps bound
half-open repetition intervals.
A clip consists of  $k$ consecutive complete repetitions from one source. Left bars mark
example windows: $k{=}1$ (pink, Rep~1), two overlapping $k{=}2$ placements
(teal, Reps~1--2; blue, Reps~2--3), and $k{=}3$ (orange, all three). Start,
middle, and end placements vary clip position while the target count stays
fixed.}
\label{fig:construction}
\end{figure}

In each source video, a repetition begins when the action leaves a rest pose
and ends when it returns to one, so consecutive rest-pose timestamps define
each repetition as a half-open interval $[\text{start},\text{end})$. We discard
zero-length and non-monotonic intervals and concatenate $k\in\{1,\ldots,20\}$
consecutive repetitions into a clip (Figure~\ref{fig:construction}). Because
adjacent repetitions share a rest-pose boundary, every clip contains only
complete repetitions and ends at a rest pose. To vary clip position while
holding the action and source fixed, we extract windows from the start,
middle, and end of the source when it contains enough repetitions; redundant
middle windows are omitted for short sources. For a given source and $k$,
these clips contain the same number of repetitions at different positions in
the source; their durations can differ because action speed varies within a
source.

\begin{figure}[!ht]
\centering
\includegraphics[width=\linewidth]{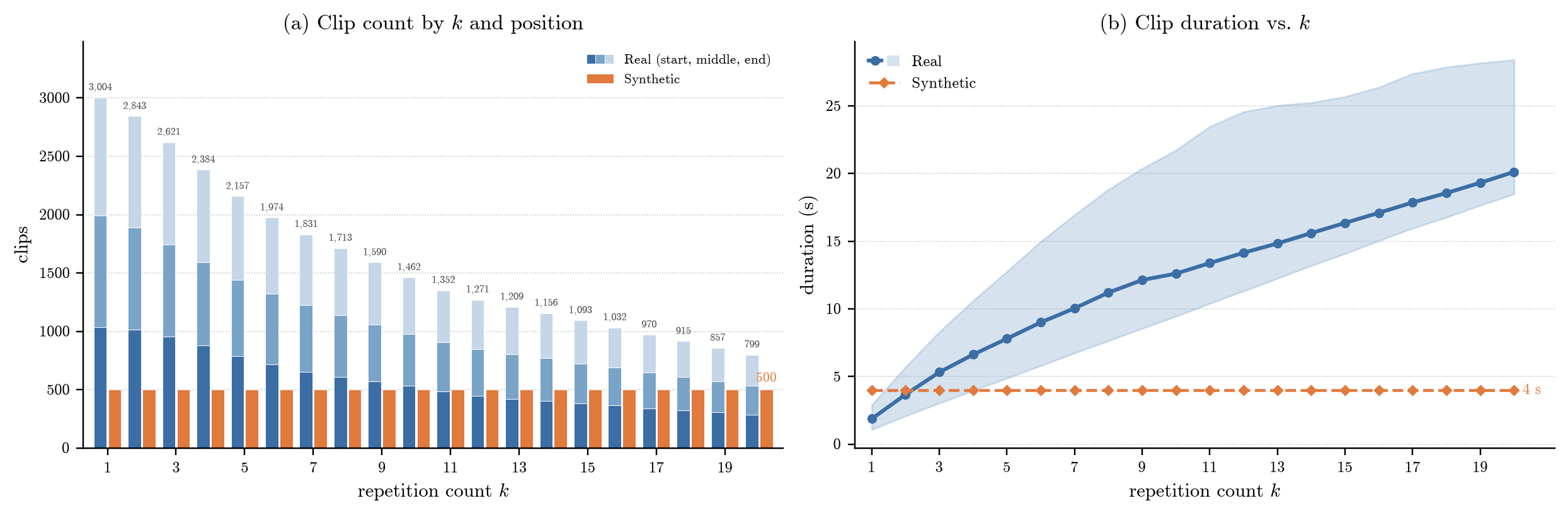}
\caption{\textbf{RepProbe statistics.} \textbf{(a)}~Clips per repetition count
$k$. RepProbe-Real (blue, stacked by clip position: start, middle, end) is
skewed toward small counts because fewer source videos contain many
repetitions; RepProbe-Synthetic (orange) is balanced at $500$ clips per count.
\textbf{(b)}~Clip duration versus $k$. Real clip duration grows with $k$ (line:
median; shaded band: spread across clips), whereas every synthetic clip lasts
$4$\,s regardless of $k$.}
\label{fig:stats}
\end{figure}

\paragraph{Statistics.}
RepProbe-Real contains $32{,}233$ clips from $1{,}038$ RepCount source videos. All splits
are made at the source-video level, so clips from the same source never cross a
split boundary. Because fewer source videos contain many annotated repetitions,
the eligible pool shrinks as $k$ grows, and the full set is skewed toward
smaller counts (Figure~\ref{fig:stats}). We use the full set for the main
evaluation and construct nested $k$-balanced subsets, stratified by clip
position, for controlled analyses: a $2{,}000$-clip subset ($100$ per count) and
a $10{,}000$-clip subset ($500$ per count). Unless otherwise stated, controlled
analyses on real video use the balanced $10$K subset. 

\subsection{RepProbe-Synthetic}
\label{sec:dataset-synthetic}

Real clips confound the periodic signal with object semantics, scene
appearance, and camera variation such as viewpoint and
lighting, so a model or probe could count by exploiting content cues
rather than periodicity itself. To isolate periodicity, RepProbe-Synthetic
contains $10{,}000$ procedurally generated clips in a
controlled setting: the set is balanced across counts ($500$ clips for each
$k=1$--$20$), and in every clip a single attribute changes periodically and
completes exactly $k$ periods, while everything else stays fixed.

\paragraph{Construction.}
Each clip shows simple geometric elements on a plain background (except for
the Textured flash stress condition), and exactly one attribute changes
periodically. The count is therefore defined only by how
many periods the attribute completes: no single frame reveals it, and there is
no object category or action that could evoke a typical count. Every clip lasts
$4$\,s regardless of $k$ (Figure~\ref{fig:stats}b), so, unlike in real video,
clip duration carries no information about the count. 

\begin{figure}[!ht]
\centering
\includegraphics[width=\linewidth]{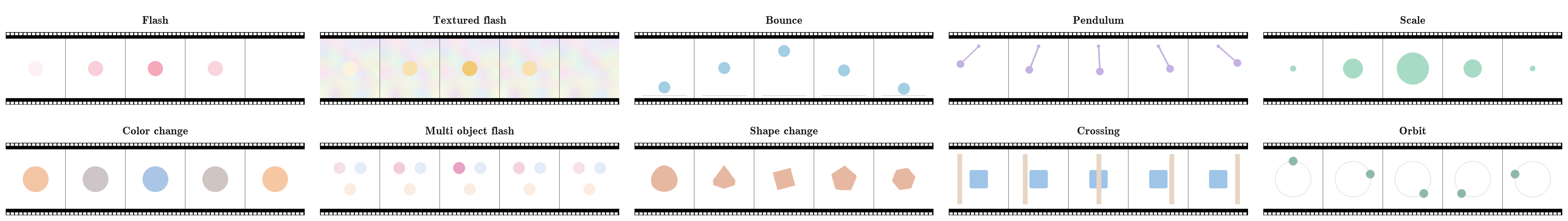}
\caption{\textbf{RepProbe-Synthetic families.} Each panel shows five frames
spanning one period of one of the ten families. In each family, a single
attribute (visibility, size, hue, position, angle, geometry, or occlusion)
changes periodically and completes $k$ periods per clip, while everything else stays fixed. Textured flash and Multi object flash add stress conditions: a
cluttered background, and flashes distributed across several objects that
take turns.}
\label{fig:synth}
\end{figure}

The ten families vary what changes periodically (Figure~\ref{fig:synth}) and
fall into three groups. Appearance changes alter an element in place: a
disc appears and disappears (Flash), pulses in size (Scale), cycles in hue
(Color change), or morphs through polygons (Shape change). Motion
changes move an element along a path: a bob hops vertically (Bounce), swings
on a pivot (Pendulum), or travels around a ring (Orbit), and a bar sweeps
across a square (Crossing). Stress conditions make the periodic signal
harder to isolate: Textured flash places the same flash on a cluttered,
scene-like background, and Multi object flash distributes the flashes across
several objects that take turns, so that every flash must be counted
regardless of where it occurs. Together, the families test whether a model
tracks periodic change in general or only specific kinds of change, such as
motion.

\subsection{Task, Metrics, and Evaluation Controls}
\label{sec:dataset-task}

\paragraph{Task and metrics.}
Each model is queried with the instruction ``\textit{Count the number of
repetitions in this video. Return only an integer.}'' A shared, model-agnostic
parser extracts the predicted count $\hat{k}$ from each response
(App.~\ref{app:statistics}). Responses from which no integer can be extracted
are treated as parse failures and scored as a prediction of $0$, so a model
cannot improve its score by failing to answer. For the main comparison across
models, predictions are clipped to $[0,40]$ so that rare extreme outputs do not
dominate mean errors. We report mean absolute error (MAE), which is directly
interpretable in repetitions, and normalized MAE ($\mathrm{MAE}_n$), which we
use when comparing across count ranges:
\[
\mathrm{MAE}=\frac{1}{N}\sum_i |\hat{k}_i-k_i|,
\qquad
\mathrm{MAE}_n=\frac{1}{N}\sum_i \frac{|\hat{k}_i-k_i|}{k_i}.
\]
We also report exact-match accuracy and off-by-one accuracy (OBO), the
percentage of predictions with $|\hat{k}-k|\leq1$. OBO is standard in
repetition counting \citep{dwibedi2020counting,hu2022transrac}; it tolerates
ambiguity at repetition boundaries while still penalizing genuine miscounts.
Results are further broken down by count band: Few ($k=1$--$6$), Mid
($k=7$--$13$), and Many ($k=14$--$20$).

\paragraph{Evaluation controls.}
Unless otherwise stated, each model receives $16$ frames per clip, sampled
uniformly regardless of duration. Cosmos-Reason2 is the exception: it uses its
native $4$\,fps setting, giving a median of $57$ frames per RepProbe-Real clip.
Cross-model differences in the main comparison therefore partly reflect frame
budget. Our central comparison, between each model's answer and a probe on its
own representation (Section~\ref{sec:exp-readout}), uses the same input for
both readouts and is unaffected. RepProbe-Real also admits potential
shortcuts: clip duration is correlated with $k$, and fewer source videos are
eligible at high counts. Source-level splits prevent leakage, and clip position sampling balances clips from the start, middle, and end of each source video.
Remaining effects of duration, clip position, and temporal coverage are
analyzed in Appendices~\ref{app:fpc} and~\ref{app:duration}.
\section{Experiments}
\label{sec:experiments}

We organize our experiments around three questions. \textbf{(1)} How well do
video language models count repetitions, and does performance hold as the
number of repetitions grows? (Section~\ref{sec:exp-main}) \textbf{(2)} When
counting fails, is the count absent from the model's internal state, or
present but not expressed in the generated answer? (Section~\ref{sec:exp-readout})
\textbf{(3)} Does the decodable count capture the actual repetitions, or does it rely on simpler cues such as video duration or scene content? 
(Section~\ref{sec:exp-signal})

\paragraph{Evaluated Models.} 
We evaluate two specialized repetition-counting models, ESCounts
\citep{sinha2024every} and RepNet \citep{dwibedi2020counting}, and seven
general-purpose video language models spanning multiple families and architectures:
Cosmos-Reason2-8B~\citep{azzolini2025cosmos}, Qwen3-VL-8B~\citep{bai2025qwen3},
Video-R1-7B~\citep{feng2026video}, VideoChat-R1.5-7B~\citep{yan2026videochat},
InternVL3.5-8B~\citep{wang2025internvl3}, LongVILA-R1-7B~\citep{chen2026scaling},
and LLaVA-OneVision-7B~\citep{li2024llava}. The specialists are trained for
repetition counting; the generalists are evaluated zero-shot.

\subsection{Behavioral Accuracy: A Collapse at High Counts}
\label{sec:exp-main}

We evaluate specialist repetition-counting models and generalist video language
models on the full RepProbe-Real set ($\sim$32K clips), which spans $k=1$ to
$20$ and is weighted toward smaller counts because fewer source videos contain
enough repetitions to support large $k$; the controlled analyses that follow
use count-balanced subsets. The specialist ESCounts remains accurate across the full range: it reaches
$\mathrm{MAE}_n=0.19$ and $41.9\%$ exact accuracy, and retains $65.6\%$ OBO
even in the Many band. In contrast, every generalist falls
to at most $22.2\%$ OBO in the Many band (Table~\ref{tab:main}). The two
strongest generalists, Cosmos-Reason2-8B and Qwen3-VL-8B, count reasonably well
at low counts ($62.1\%$ and $62.9\%$ OBO in the Few band) but reach only
$22.2\%$ and $4.9\%$ OBO in the Many band, and the remaining generalists stay
below $8\%$ Many-band OBO.

\begin{table}[t]
\centering
\small
\setlength{\tabcolsep}{3pt}
\caption{Repetition counting on RepProbe-Real and RepProbe-Synthetic
(generalists only), $k=1$--$20$. Predictions are clipped to $[0,40]$. Bands:
Few $k{=}1$--$6$, Mid $k{=}7$--$13$, Many $k{=}14$--$20$. \textbf{Bold}: best
generalist in each column. The shaded Many band highlights the high-count
collapse on RepProbe-Real. }
\label{tab:main}
\begin{tabular}{@{}l cccc cc>{\columncolor{gray!12}}c@{}}
\toprule
& \multicolumn{4}{c}{Overall} & \multicolumn{3}{c}{OBO (\%) by band $\uparrow$} \\
\cmidrule(lr){2-5}\cmidrule(l){6-8}
Method & MAE$_n\downarrow$ & MAE$\downarrow$ & Exact$\uparrow$ & OBO$\uparrow$
& Few & Mid & Many \\
\midrule
\multicolumn{8}{@{}l}{\textbf{RepProbe-Real}} \\
\multicolumn{8}{@{}l}{\quad\emph{Specialist}} \\
ESCounts & 0.19 & 1.07 & 41.9 & 77.5 & 87.5 & 71.0 & 65.6 \\
RepNet   & 0.32 & 2.20 & 21.7 & 53.0 & 79.1 & 36.4 & 21.3 \\
\cmidrule(l){1-8}
\multicolumn{8}{@{}l}{\quad\emph{Generalist}} \\
Cosmos-Reason2-8B  & \textbf{0.43} & \textbf{2.89} & \textbf{17.5} & \textbf{47.5} & 62.1 & \textbf{43.1} & \textbf{22.2} \\
Qwen3-VL-8B        & 0.68 & 5.02 & 12.9 & 35.0 & \textbf{62.9} & 14.6 & 4.9 \\
Video-R1-7B        & 1.35 & 6.24 & 7.2 & 19.3 & 24.8 & 21.4 & 4.2 \\
VideoChat-R1.5-7B  & 1.40 & 6.41 & 7.0 & 19.4 & 25.7 & 21.0 & 3.2 \\
LongVILA-R1-7B     & 1.47 & 7.70 & 6.5 & 18.4 & 33.1 & 9.3 & 2.2 \\
InternVL3.5-8B     & 1.51 & 5.98 & 7.3 & 18.7 & 21.0 & 23.0 & 7.3 \\
LLaVA-OneVision-7B & 2.19 & 6.80 & 4.8 & 12.8 & 2.1 & 33.0 & 5.3 \\
\midrule
\multicolumn{8}{@{}l}{\textbf{RepProbe-Synthetic}} \\
Cosmos-Reason2-8B  & 0.71 & 7.75 & 5.6 & 16.1 & 53.7 & 0.0 & 0.0 \\
Qwen3-VL-8B        & \textbf{0.69} & 7.77 & \textbf{6.3} & \textbf{17.5} & \textbf{58.3} & 0.0 & 0.0 \\
Video-R1-7B        & 0.73 & 8.37 & 4.7 & 15.3 & 50.9 & 0.0 & 0.0 \\
VideoChat-R1.5-7B  & 0.72 & 7.96 & 5.2 & 14.8 & 49.1 & 0.0 & 0.0 \\
LongVILA-R1-7B     & 0.77 & 8.43 & 4.4 & 13.6 & 44.3 & 0.8 & 0.0 \\
InternVL3.5-8B     & 0.70 & \textbf{6.50} & 5.1 & 16.3 & 44.9 & 8.0 & 0.0 \\
LLaVA-OneVision-7B & 0.87 & 7.13 & 5.6 & 13.5 & 29.5 & \textbf{13.3} & 0.0 \\
\bottomrule
\end{tabular}
\end{table}

\paragraph{The collapse is not caused by sparse frame sampling.}
Under a fixed frame budget, larger $k$ leaves fewer sampled frames per
repetition. We therefore fix the number of frames per cycle (FPC) and vary $k$
independently. High counts remain harder even at the densest setting we test:
at FPC$\,=8$, the MAE of Qwen3-VL-8B rises from $2.93$ at $k=5$ to $7.86$ at
$k=20$, and that of Cosmos-Reason2-8B from $2.47$ to $4.57$. Coverage does
matter at fixed $k$: for Cosmos-Reason2-8B at $k=20$, increasing coverage from
one to eight frames per cycle reduces MAE from $14.00$ to $4.57$. Coverage and
count therefore have separable effects, and the effect of count remains at
every coverage level we test. Temporal coverage affects accuracy but does not
account for the collapse.

\paragraph{The weakness is not specific to exercise video.}
On RepProbe-Synthetic, in which only a single attribute changes periodically,
generalists struggle even at low counts, reaching only $29.5\%$ to $58.3\%$
OBO in the Few band. We evaluate only the generalists here, since the
specialists are trained on real exercise video and these stimuli fall outside
their training domain. On RepProbe-Synthetic, the relevant comparison is
instead between each generalist's answer and a linear probe on its own
internal state (Section~\ref{sec:exp-readout}).

These results establish a behavioral failure but not its source: the count may
be poorly represented inside the model, or represented but not converted into
the numerical answer. The full FPC grid is given in Appendix~\ref{app:fpc}.

\subsection{Representation vs.\ Output: Decodable but Poorly Reported}
\label{sec:exp-readout}

The behavioral results show that the model produces the wrong count, but not
whether the count is missing from its representation or present but lost on
the way to the answer. To separate these, we read the count in two ways from
the same final-layer state, the one the language head consumes to produce its
answer. The first readout is the model's answer: the number it generates when
asked. The second is a count probe. We extract the residual-stream activation after
the final transformer block at the last prompt token, the position from which
the language head produces the first answer token, and train a linear
classifier over the $20$ count classes ($k=1,\dots,20$) on $1{,}000$ labeled
clips. The probe's prediction is the most probable class. On RepProbe-Real,
training and evaluation clips are split at the source-video level, so the
probe is always evaluated on source videos it has not seen during training;
full details are given in Appendix~\ref{app:additional}.
Because the probe is linear, it
cannot perform nonlinear computation on the state; it can only recover
information that is already linearly present. The language head is itself a
linear map applied to this same state (after normalization), so the probe has
no more capacity than the model's own readout: any count the probe recovers is
available to the language head in the form it consumes. A gap between the two
readouts therefore locates the failure in the output pathway rather than in
the representation. Both readouts are computed from the same forward pass on
the same frames, so the comparison is not confounded by differences in input.
We study Cosmos-Reason2-8B and Qwen3-VL-8B, the two strongest generalists
(Table~\ref{tab:main}). Because the probe is trained with supervision, we also
test in Section~\ref{sec:exp-signal} whether it reads the count itself or
correlates of it, such as duration, clip position, or scene content.

\begin{figure}[t]
    \centering
    \includegraphics[width=\linewidth]{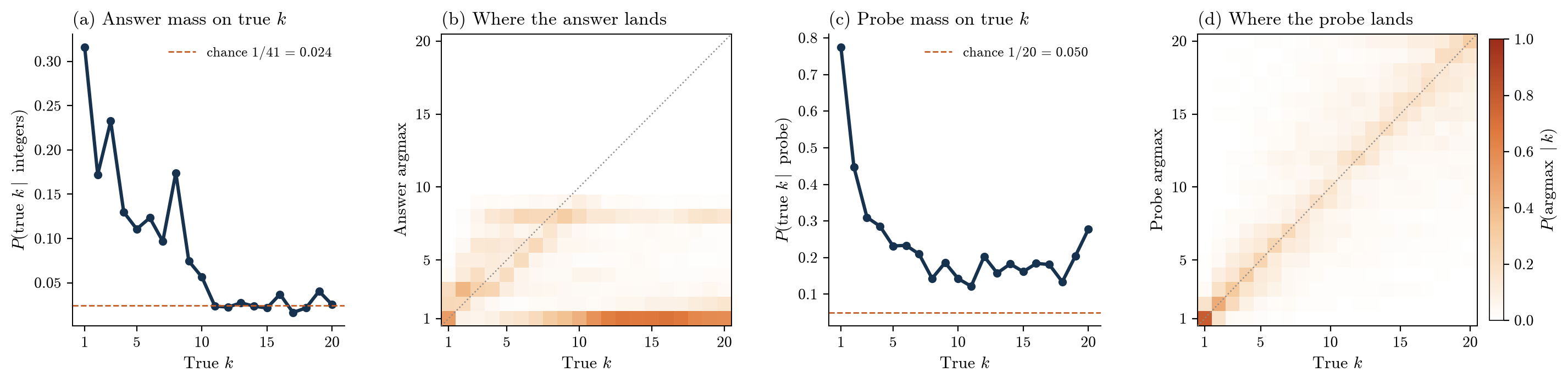}
    \caption{\textbf{The count is recoverable by a probe but not by the model's
answer} (Cosmos-Reason2, RepProbe-Real). \textbf{(a)}~The answer's probability
on the true $k$ falls to chance by $k\!\approx\!11$; \textbf{(b)}~its argmax
collapses onto a few small numbers. \textbf{(c)}~The probe keeps far more mass
on the true $k$; \textbf{(d)}~its argmax tracks $k$ across the full range.}
    \label{fig:number-prior}
\end{figure}

The probe recovers the count far more accurately than the model's answer
(Table~\ref{tab:fairness-both}, Figure~\ref{fig:number-prior}). On
RepProbe-Real, the probe lowers MAE from $3.91$ to $2.49$ for Cosmos-Reason2
and from $7.88$ to $3.05$ for Qwen3-VL. As a control, a Cosmos-Reason2 probe trained on the same clips with shuffled
labels performs no better than random guessing (MAE $6.57$). The gap concentrates at high $k$: in the Many band, the probe reaches
MAE $2.75$ against $6.35$ for Cosmos-Reason2 and $3.54$ against $13.43$ for
Qwen3-VL. Across paired clips, the probe improves on the Cosmos-Reason2 answer
by $1.42$ MAE on average (95\% CI $[1.29, 1.54]$). The dissociation also holds
on RepProbe-Synthetic: in the Few band, the probe reaches $92.3\%$ and $93.9\%$
OBO for Cosmos-Reason2 and Qwen3-VL, while their answers reach $53.8\%$ and
$58.4\%$. In both settings, the count is read far more accurately from the
representation than from the model's own answer.

\begin{table}[t]
\centering
\small
\setlength{\tabcolsep}{2.3pt}
\caption{Two ways of reading the count from the same final-layer representation,
on the balanced $10$K subset of RepProbe-Real and on RepProbe-Synthetic. The
linear count probe (trained on 1k clips) is more accurate than the model's own
answer on both; on RepProbe-Real, the gap is largest at high $k$.}
\label{tab:fairness-both}
\begin{tabular}{@{}ll*{8}{r}@{}}
\toprule
&& \multicolumn{2}{c}{All} & \multicolumn{2}{c}{Few} &
\multicolumn{2}{c}{Mid} & \multicolumn{2}{c}{Many}\\
\cmidrule(lr){3-4}\cmidrule(lr){5-6}\cmidrule(lr){7-8}\cmidrule(lr){9-10}
Model & Readout &
\multicolumn{1}{c}{MAE$\downarrow$} & \multicolumn{1}{c}{OBO$\uparrow$} &
\multicolumn{1}{c}{MAE$\downarrow$} & \multicolumn{1}{c}{OBO$\uparrow$} &
\multicolumn{1}{c}{MAE$\downarrow$} & \multicolumn{1}{c}{OBO$\uparrow$} &
\multicolumn{1}{c}{MAE$\downarrow$} & \multicolumn{1}{c}{OBO$\uparrow$}\\
\midrule
\multicolumn{10}{@{}l}{\textbf{\textit{RepProbe-Real}}}\\[3pt]
Cosmos-8B & Model
& 3.91 & 37.9 & 1.91 & 56.0 & 3.20 & \textbf{40.7} & 6.35 & 19.4\\
& Probe
& \textbf{2.49} & \textbf{50.3} & \textbf{1.62} & \textbf{68.0}
& \textbf{2.99} & 39.7 & \textbf{2.75} & \textbf{45.6}\\
Qwen3-VL-8B & Model
& 7.88 & 24.2 & 2.45 & \textbf{58.9} & 6.97 & 14.0 & 13.43 & 4.7\\
& Probe
& \textbf{3.05} & \textbf{44.7} & \textbf{2.27} & \textbf{58.9}
& \textbf{3.23} & \textbf{37.9} & \textbf{3.54} & \textbf{39.4}\\
\midrule
\multicolumn{10}{@{}l}{\textbf{\textit{RepProbe-Synthetic}}}\\[3pt]
Cosmos-8B & Model
& 7.75 & 16.1 & 1.51 & 53.8 & 6.90 & 0.0 & 13.95 & 0.0\\
& Probe
& \textbf{0.62} & \textbf{91.4} & \textbf{0.59} & \textbf{92.3}
& \textbf{0.62} & \textbf{90.7} & \textbf{0.65} & \textbf{91.3}\\
Qwen3-VL-8B & Model
& 7.77 & 17.5 & 1.50 & 58.4 & 6.99 & 0.0 & 13.91 & 0.0\\
& Probe
& \textbf{0.59} & \textbf{92.5} & \textbf{0.50} & \textbf{93.9}
& \textbf{0.66} & \textbf{90.5} & \textbf{0.58} & \textbf{93.5}\\
\bottomrule
\end{tabular}
\end{table}

\paragraph{The count is decodable before the final layer.}
Probes trained at earlier layers show that the count is already linearly
decodable at intermediate depth and remains available at the final block. At
block $18$, probe MAE is $2.97$ for Qwen3-VL and $2.51$ for Cosmos-Reason2,
compared with $3.05$ and $2.49$ at the final block (App.~\ref{app:additional}).
The count is therefore not lost as information propagates toward the output;
it is available at the point where the language head reads the state.

\paragraph{The output follows a number prior.}
The model's answers concentrate on a small set of numbers regardless of the
true count (Figure~\ref{fig:number-prior}b). Because two-digit numbers span
several digit tokens, we also compare the single-token number words ``one''
through ``twenty'' (a length-fair readout). Under this readout, the output
settles on a single preferred word (``fifteen'' for Qwen3-VL, ``eighteen'' for
Cosmos-Reason2), and across answers, how often each number is produced tracks
its lexical frequency in text (Pearson $r=0.41$). The count is therefore not
absent: it is linearly available in the final internal state, and the failure
lies in converting it into an answer, which instead follows a small set of
preferred numbers.

% ------------------------------------------------------------
% 4.3
% ------------------------------------------------------------
\subsection{Controls: The Decodable Count Indeed Reflects the Repetitions}
\label{sec:exp-signal}

Section~\ref{sec:exp-readout} shows that the count is linearly decodable from
the final-layer representation of RepProbe-Real clips even at high $k$, where
the model's own answer collapses. Decodability alone, however, does not show
what the probe relies on: it could exploit correlates of the count rather than
the count itself. We test two such alternatives.

\paragraph{Duration and clip position.}
Count and duration correlate over RepProbe-Real ($r=0.62$) but only weakly at
high $k$ ($r=0.12$), and the probe's advantage over the answer survives
controlling for duration: for Cosmos-Reason2, a duration-only predictor
reaches normalized MAE $0.57$ against $0.30$ for the probe, and after
partialling out duration, the probe's predictions correlate with the true
count at $0.71$ against $0.33$ for the answer. On RepProbe-Synthetic, every
clip lasts $4$\,s, so duration carries no information about the count at all.
A separate probe for the clip's position within its source video (start,
middle, or end) reaches only $38.7\%$ accuracy against $33.3\%$ chance, and
this accuracy stays flat across $k$ (Appendix~\ref{app:duration}). Position is
therefore only weakly encoded and cannot account for a count signal that
varies with $k$.

\paragraph{Scene content.}
The probe could also read a scene-level correlate of the count in exercise
video. RepProbe-Synthetic rules this out: its clips contain only a single periodically changing attribute, across ten families
that vary what changes, clutter, and background. The count is defined only by
how many periods the attribute completes and cannot be read from any single
frame. On these stimuli, the probe still recovers the count far more accurately
than the model's answer (Section~\ref{sec:exp-readout}), so the dissociation
holds when the periodic change is the only available signal. The decodable count therefore reflects repetition structure rather than
duration, clip position, or scene content. Together with
Section~\ref{sec:exp-readout}, this characterizes the high-count failure as a
representation-to-output gap: the count is linearly available in the final representation, but the output pathway does not express it, instead favoring a small set of preferred numbers.
\section{Conclusion}
\label{sec:conclusion}

We introduced RepProbe, a controlled benchmark for studying repetition
counting in video language models, comprising 42K real and synthetic clips
with exact repetition counts and controlled clip properties. Using RepProbe,
we uncovered a systematic high-count collapse in general-purpose video
language models that persists even when temporal coverage is controlled,
while specialist counting models remain accurate. Most importantly, we showed
that in the models we probe, the repetition count is often more accurately
recoverable from the final representation than from the generated answer, and
that this dissociation is not explained by duration, clip position, or scene
content. Together, these findings suggest that repetition-counting failures
arise primarily from the output pathway rather than from an inability to
perceive or encode repetitions. More broadly, our results highlight the
importance of distinguishing representation from readout when evaluating
temporal reasoning in video language models, and suggest that improving output
mechanisms may be as important as improving visual perception for reliable
counting.

\paragraph{Limitations.}
We focus our probing analysis on the two strongest generalists, Cosmos-Reason2
and Qwen3-VL, and on repetition counting, a task that allows exact control of
the target quantity; extending the analysis to more models and to other
temporal quantities is a natural next step. RepProbe-Synthetic isolates
periodicity by design, and broadening the real-video component beyond exercise
footage would further test generality. Finally, we characterize the output-side
failure through the answers' concentration on preferred numbers; identifying the
mechanism that produces these preferences is future work.

% \newpage
% \section*{Use of Generative AI}

% We used generative AI tools to assist with writing, including editing,
% restructuring, and clarifying author-written text. All research ideas, benchmark
% construction, experiments, and results are the authors' own, and all
% analyses were run by the authors. We reviewed all model-assisted edits and
% take full responsibility for the content of this work.

% \section*{Reproducibility statement}

% The code and clip-level annotations for both the real and synthetic videos are included in the supplementary material. The videos are omitted due to their size exceeding the supplementary file-size limit. All benchmark videos and code will be made publicly available upon acceptance.

\bibliography{main}
\bibliographystyle{iclr2027_conference}

\clearpage
\appendix

\setcounter{table}{0}
\setcounter{figure}{0}
\section{Additional Experimental Details and Analyses}
\label{sec:appendix}

\setcounter{table}{0}
\setcounter{figure}{0}
\counterwithin{table}{section}
\counterwithin{figure}{section}

This appendix provides additional experimental details and robustness checks
supporting the main results. We report the hold-$k$ frames-per-cycle grid
(App.~\ref{app:fpc}), additional probe details (App.~\ref{app:additional}), the
numerical-output analysis (App.~\ref{app:numprior}), statistical and parsing
procedures (App.~\ref{app:statistics}), and the duration and clip-position
audit (App.~\ref{app:duration}).

% ============================================================
% Section 4.1: FPC grid
% ============================================================

\subsection{Hold-\texorpdfstring{$k$}{k} Frames-per-Cycle Grid}
\label{app:fpc}

At a fixed input frame budget, the number of frames available per repetition
decreases as $k$ grows. To separate repetition count from temporal coverage on
RepProbe-Real, we fix frames per cycle (FPC) and vary $k$ independently on a
$4\times4$ grid with $100$ clips per cell (Table~\ref{tab:fpc}). Reading down a
column isolates the effect of coverage at fixed $k$; reading across a row
isolates the effect of $k$ at fixed coverage.

Error rises with $k$ even at the densest coverage (FPC$\,=8$): Cosmos-Reason2
increases from MAE $2.47$ at $k=5$ to $4.57$ at $k=20$, and Qwen3-VL from
$2.93$ to $7.86$. At fixed $k$, additional coverage can substantially reduce
error; for example, Cosmos-Reason2 at $k=20$ improves from MAE $14.00$ at
FPC$\,=1$ to $4.57$ at FPC$\,=8$. Count and coverage therefore have separable
effects, and coverage alone does not remove the high-$k$ degradation. With
$100$ clips per cell, individual cells are noisy, particularly for Qwen3-VL at
high $k$; the conclusion rests on the overall trend rather than on any single
cell.

\begin{table}[ht!]
\centering\small
\caption{Per-cell MAE on the hold-$k$ $\times$ FPC grid (RepProbe-Real),
$100$ clips per cell.}
\label{tab:fpc}
\begin{tabular}{@{}llcccc@{}}
\toprule
Model & FPC & $k{=}5$ & $k{=}10$ & $k{=}15$ & $k{=}20$ \\
\midrule
\multirow{4}{*}{Qwen3-VL-8B}
& 8 & 2.93 & 4.09 & 7.98 & 7.86 \\
& 4 & 3.50 & 4.83 & 6.82 & 14.18 \\
& 2 & 2.96 & 5.16 & 13.64 & 21.42 \\
& 1 & 3.90 & 6.13 & 9.25 & 14.20 \\
\midrule
\multirow{4}{*}{Cosmos-Reason2-8B}
& 8 & 2.47 & 2.13 & 4.46 & 4.57 \\
& 4 & 2.19 & 3.14 & 4.17 & 6.16 \\
& 2 & 2.11 & 3.98 & 6.61 & 10.62 \\
& 1 & 3.20 & 4.70 & 8.59 & 14.00 \\
\bottomrule
\end{tabular}
\end{table}

% ============================================================
% Section 4.2: probe details, numerical output, statistics
% ============================================================

\subsection{Additional Probe Details}
\label{app:additional}

\paragraph{Probe.}
For each clip, we run a single forward pass on the prompt and video and record
the residual-stream activation after the final transformer block at the last
prompt token, the position from which the language head produces the first
answer token. The count probe is a multinomial logistic-regression classifier
over the $20$ count classes ($k=1,\dots,20$), trained on these activations for
$1{,}000$ labeled clips; its prediction is the most probable class, and its
class probabilities are those shown in Figure~\ref{fig:number-prior}c. On
RepProbe-Real, training and evaluation clips are split at the source-video
level, so that no source video contributes clips to both. Probe activations
and the model's generated answer come from the same forward pass on the same
input frames. To see how count decodability changes with depth, we also train
probes at earlier layers. All layers are probed with the same training clips,
evaluation clips, and classifier. At the outputs of blocks $0$, $18$, and $35$,
probe MAE is $2.91$, $2.51$, and $2.49$ for Cosmos-Reason2, and $4.09$, $2.97$,
and $3.05$ for Qwen3-VL; the block-$35$ values match
Table~\ref{tab:fairness-both}. These results indicate that repetition-count
information becomes substantially more decodable in deeper layers, with most
of the improvement occurring before the final block; however, the near-flat
MAE for Cosmos-Reason2 and the slight degradation at the final layer for
Qwen3-VL suggest that later layers do not necessarily make the count signal
more linearly decodable.

\paragraph{Paired evaluation set.}
On RepProbe-Real, the probe and the generated answer are compared on the clips
of the balanced $10$K subset for which both readouts are available:
$9{,}950$ clips for Cosmos-Reason2 and $9{,}989$ for Qwen3-VL.

\subsection{Detailed Numerical-Output Analysis}
\label{app:numprior}

Figure~\ref{fig:number-prior} shows the distributions underlying the
numerical-output analysis in Section~\ref{sec:exp-readout}, computed on
RepProbe-Real. Reading the count directly from digit tokens is biased, because
single-digit integers are one token while two-digit integers span several
tokens, and the first token of $10$ through $19$ coincides with the digit $1$.
We therefore use a length-fair word readout that compares the single-token
number words ``one'' through ``twenty.'' This readout can select two-digit
values, yet predictions still concentrate on a preferred number: the mode is
``fifteen'' for Qwen3-VL and ``eighteen'' for Cosmos-Reason2 across the count
range. The concentration is therefore not explained by token-length asymmetry.

Freely generated counts also correlate with lexical number-word frequency
($r=0.41$ over the $1$--$20$ count histogram), and salient numbers are
overrepresented, including $10$ at $13.7\%$ and $12$ at $7.6\%$ of answers.

\subsection{Statistical and Parsing Details}
\label{app:statistics}

Paired bootstrap confidence intervals are computed by resampling evaluation
clips with replacement and recomputing the paired difference in MAE for each
of $2{,}000$ replicates. Resampling is performed at the clip level because
paired comparisons evaluate the same clips under different readouts.

Generated responses are parsed with a deterministic numerical parser. If an
\texttt{<answer>...</answer>} span is present, we extract the last integer inside
it. Otherwise, we scan upward from the final line and select the first line
consisting solely of an integer after removing trailing punctuation. If neither
rule succeeds, we use the final match of \texttt{-?\textbackslash d+} in the
response. No separate chain-of-thought stripping is applied. Responses without
an interpretable count are treated as parse failures and scored as a prediction of $0$. On the balanced $10$K subset of RepProbe-Real, failure
rates are at most $0.11\%$ for Cosmos-Reason2, Qwen3-VL, InternVL3.5, Video-R1,
and VideoChat-R1.5, $0.67\%$ for LLaVA-OneVision, and $1.50\%$ for
LongVILA-R1. Responses that were not produced at all are excluded.

% ============================================================
% Section 4.3: duration, clip position
% ============================================================

\subsection{Duration and Clip-Position Audit}
\label{app:duration}

Clip duration is correlated with repetition count and therefore provides a
potential shortcut for predicting $k$. We audit this on the full RepProbe-Real
set and on its balanced $10{,}000$-clip subset. The Pearson correlation between
duration and count is $r=0.62$ and $0.55$, respectively, and the Spearman
correlation is $\rho=0.75$ and $0.64$; the corresponding linear-regression
$R^2$ values are $0.39$ and $0.30$. On RepProbe-Synthetic, every clip lasts
$4$\,s, so duration carries no information about the count.

The association on RepProbe-Real weakens at larger counts: Pearson $r$
decreases from $0.50$ in the Few band to $0.21$ in the Mid band and $0.12$ in
the Many band. Duration thus carries information about $k$, but much less in
the high-count regime.

For Cosmos-Reason2, a duration-only isotonic predictor reaches normalized MAE
$0.57$, compared with $0.30$ for the probe. After controlling for duration, the
partial correlation between predicted and true count is $0.71$ for the probe
and $0.33$ for the generated answer. Within $2$-second duration bins, probe MAE
remains lower than answer MAE ($2.87$ versus $4.59$). Duration alone therefore
does not account for the gap between the probe and the answer.

\begin{table}[!t]
\centering
\small
\setlength{\tabcolsep}{4pt}
\caption{Counting accuracy by clip position on the balanced $10$k RepProbe-Real set
($\approx 3{,}330$ clips in each window). Parse failures scored as $0$, then
clipped to $[0,40]$.}
\label{tab:position}
\begin{tabular}{@{}l ccc @{\hspace{8pt}} ccc@{}}\toprule
 & \multicolumn{3}{c}{Exact\,$\uparrow$} & \multicolumn{3}{c}{OBO\,$\uparrow$} \\
\cmidrule(r){2-4}\cmidrule(l){5-7}
Method & start & middle & end & start & middle & end \\
\midrule
ESCounts & \num{37.6} & \num{36.9} & \num{37.0} & \num{72.5} & \num{72.3} & \num{70.7} \\
RepNet & \num{16.5} & \num{17.1} & \num{16.7} & \num{42.3} & \num{43.0} & \num{42.9} \\
Cosmos probe (L35) & \num{24.1} & \num{26.0} & \num{24.8} & \num{49.2} & \num{51.7} & \num{49.8} \\
Qwen probe (L35) & \num{21.4} & \num{23.5} & \num{21.3} & \num{43.7} & \num{46.5} & \num{44.2} \\
Cosmos-Reason2-8B & \num{14.2} & \num{12.3} & \num{12.4} & \num{38.1} & \num{38.1} & \num{37.3} \\
Qwen3-VL-8B & \num{9.8} & \num{8.6} & \num{8.9} & \num{24.8} & \num{23.7} & \num{24.2} \\
InternVL3.5-8B & \num{8.7} & \num{5.1} & \num{5.3} & \num{19.2} & \num{15.7} & \num{16.6} \\
LLaVA-OneVision-7B & \num{7.2} & \num{4.5} & \num{4.7} & \num{16.9} & \num{13.4} & \num{14.3} \\
VideoChat-R1.5-7B & \num{6.9} & \num{5.2} & \num{4.7} & \num{17.5} & \num{14.5} & \num{14.5} \\
Video-R1-7B & \num{6.5} & \num{4.8} & \num{5.0} & \num{17.5} & \num{13.4} & \num{14.6} \\
LongVILA-R1-7B & \num{4.6} & \num{4.6} & \num{4.6} & \num{13.1} & \num{13.1} & \num{13.4} \\
\bottomrule
\end{tabular}
\end{table}

\paragraph{Clip position.}
A three-way linear classifier on the same last-block residual of Cosmos-Reason2
predicts whether a clip was cut from the start, middle, or end of its source
video. It reaches $38.7\%$ accuracy against $33.3\%$ chance, and this rate stays
approximately constant across $k$ ($34$--$44\%$). Tab.~\ref{tab:position} scores
counting itself on the same three windows ($\approx 3{,}330$ clips each). Exact
and OBO move by at most a few points (InternVL $8.7\!\to\!5.1$ Exact is the
largest shift); specialists and LongVILA are flat. Clip position is therefore
only weakly encoded and cannot account for the count signal read by the probe.

\end{document}